\documentclass[letterpaper, 10 pt, conference]{ieeeconf}  

\IEEEoverridecommandlockouts                              

\usepackage{graphics} 
\usepackage{epsfig} 
\usepackage{mathptmx} 
\usepackage{times} 
\usepackage{amsmath} 
\usepackage{amssymb}  
\usepackage{gensymb}
\usepackage{color}
\usepackage[pagebackref=true,breaklinks=true,letterpaper=true,colorlinks,bookmarks=false]{hyperref}

\title{\LARGE \bf
Target-driven Visual Navigation in Indoor Scenes\\ using Deep Reinforcement Learning$^*$}

\author{\qquad Yuke Zhu$^{1}$ \qquad Roozbeh Mottaghi$^{2}$ \qquad Eric Kolve$^{2}$ \qquad Joseph J. Lim$^{1}$ \qquad Abhinav Gupta$^{2,3}$  \\ \qquad Li Fei-Fei$^{1}$ \qquad Ali Farhadi$^{2,4}$
\\ $^{1}$Stanford University, $^{2}$Allen Institute for AI, $^{3}$Carnegie Mellon University, $^{4}$University of Washington 
\thanks{* This work is part of the Plato project of the Allen Institute for Artificial Intelligence (AI2) and it was done while the first author was an intern at AI2.}
}

\begin{document}

\maketitle
\thispagestyle{empty}
\pagestyle{empty}

\begin{abstract}
Two less addressed issues of deep reinforcement learning are (1) lack of generalization capability to new target goals, and (2) data inefficiency i.e., the model requires several (and often costly) episodes of trial and error to converge, which makes it impractical to be applied to real-world scenarios. In this paper, we address these two issues and apply our model to the task of target-driven \emph{visual} navigation. To address the first issue, we propose an actor-critic model whose policy is a function of the goal as well as the current state, which allows to better generalize. To address the second issue, we propose AI2-THOR framework, which provides an environment with high-quality 3D scenes and physics engine. Our framework enables agents to take actions and interact with objects.
Hence, we can collect a huge number of training samples efficiently. 

We show that our proposed method (1) converges faster than the state-of-the-art deep reinforcement learning methods, (2) generalizes across targets and across scenes, (3) generalizes to a real robot scenario with a small amount of fine-tuning (although the model is trained in simulation), (4) is end-to-end trainable and does not need feature engineering, feature matching between frames or 3D reconstruction of the environment. 

The supplementary video can be accessed at the following link: \url{https://youtu.be/SmBxMDiOrvs}.

\end{abstract}



\newcommand{\ti}[1]{\textit{#1}}

\section{Introduction}

Many tasks in robotics involve interactions with physical environments and objects. One of the fundamental components of such interactions is understanding the correlation and causality between actions of an agent and the changes of the environment as a result of the action. 
Since the 1970s, there have been various attempts to build a system that can understand such relationships.
Recently, with the rise of deep learning models, learning-based approaches have gained wide popularity \cite{levine16, mnih2015human}.

In this paper, we focus on the problem of navigating a space to find a given target goal using only visual input. Successful navigation requires learning relationships between actions and the environment.
This makes the task well suited to a Deep Reinforcement Learning (DRL) approach. 
However, general DRL approaches (e.g.,  \cite{mnih2015human,mnih2016asynchronous}) 
are designed to learn a policy that depends only on the current state, and the target goal is implicitly embedded in the model parameters. Hence, it is necessary to learn new model parameters for a new target. This is problematic since training DRL agents is computationally expensive.

\begin{figure}[t]
\center
\includegraphics[width=1\linewidth]{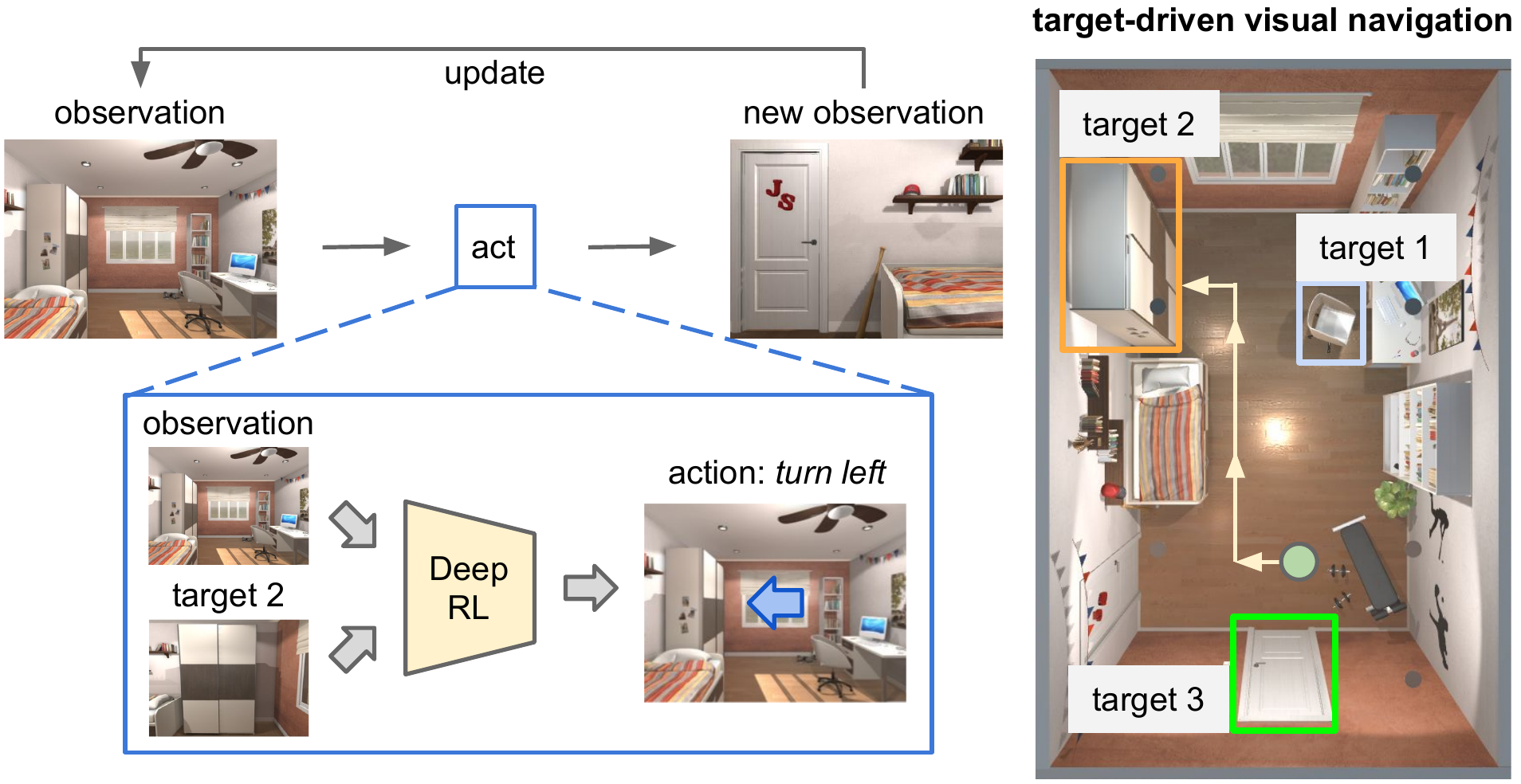}
\caption{The goal of our deep reinforcement learning model is to navigate towards a visual target with a minimum number of steps. Our model takes the current observation and the image of the target as input and generates an action in the 3D environment as the output. Our model learns to navigate to different targets in a scene without re-training.}
\vspace{-0.4cm}
\label{fig:teaser}
\end{figure}

In order to achieve higher \ti{adaptability} and \ti{flexibility}, we introduce a \ti{target-driven model}. Our model takes the visual task objective as an input, hence we can avoid re-training for every new target goal. Our model learns a policy that jointly embeds the target goal and the current state. Essentially, an agent learns to take its next action conditioned on both its current state and target, rather than its current state only. Hence, there is no need to re-train the model for new targets. A key intuition that we rely on is that different training episodes share information. For example, agents explore common routes during the training stage while being trained for finding different targets. Various scenes also share generalizable aspects (e.g., a fridge is usually near a microwave).
In short, we exploit the fact that learning for new targets will be easier with the models that have been trained for other targets.

Unfortunately, training and quantitatively evaluating DRL algorithms in real environments is often tedious. One reason is that running systems in a physical space is time consuming. Furthermore, acquiring large-scale action and interaction data of real environments is not trivial via the common image dataset collection techniques. To this end, we developed one of the first simulation frameworks with high-quality 3D scenes, called The House Of inteRactions (AI2-THOR). 
Our simulation framework enables us to collect a large number of visual observations for action and reaction in different environments. For example, an agent can freely navigate (i.e.\ move and rotate) in various realistic indoor scenes, and is able to have low- and high-level interactions with the objects (e.g.,\ applying a force or opening/closing a microwave).

We evaluate our method for the following tasks: (1) \emph{Target generalization}, where the goal is to navigate to targets that have not been used during training within a scene. (2) \emph{Scene generalization}, where the goal is to navigate to targets in scenes not used for training. (3) \emph{Real-world generalization}, where we demonstrate navigation to targets using a real robot. Our experiments show that we outperform the state-of-the-art DRL methods in terms of data efficiency for training. We also demonstrate the generalization aspect of our model. 

In summary, in this paper, we introduce a novel reinforcement learning model that generalizes across targets and scenes. To learn and evaluate reinforcement learning models, we create a simulation framework with high-quality rendering that enables visual interactions for agents. We also demonstrate real robot navigation using our model generalized to the real world with a small amount of fine-tuning.


\section{RELATED WORK}
 There is a large body of work on \emph{visual} navigation. We provide a brief overview of some of the relevant work. The map-based navigation methods require a global map of the environment to make decisions for navigation (e.g., \cite{borenstein89, borenstein91, kim95, oriolo95}). One of the main advantages of our method over these approaches is that it does not need a prior map of the environment. Another class of navigation methods reconstruct a map on the fly and use it for navigation~\cite{sim06, wooden06, davison03, tomono03}, or go through a training phase guided by humans to build the map~\cite{kinodo02,royer05}. In contrast, our method does not require a map of the environment, as it does not have any assumption on the landmarks of the environment, nor does it require a human-guided training phase. Map-less navigation methods are common as well~\cite{haddad98,lenser03,remazeilles04,saeedi06}. These methods mainly focus on obstacle avoidance given the input image. Our method is considered \emph{map-less}. However, it possesses implicit knowledge of the environment. A survey of visual navigation methods can be found in~\cite{bonin08}. 

Note that our approach is not based on feature matching or 3D reconstruction, unlike e.g.,~\cite{konolige10,phillips16}. Besides, our approach does not require supervised training for recognizing distinctive landmarks, unlike e.g., ~\cite{mcmanus14, linegar16}. 

Reinforcement Learning (RL) has been used in a variety of applications. \cite{kohl04} propose a policy gradient RL approach for locomotion of a four-legged robot. \cite{peters08} discuss policy gradient methods for learning motor primitives. \cite{michels05} propose an RL-based method for obstacle detection using a monocular camera. \cite{kim04} apply reinforcement learning to autonomous helicopter flight. \cite{kollar08} use RL to automate data collection process for mapping. \cite{barreto16} propose a kernel-based reinforcement learning algorithm for large-scale settings. \cite{liang16} use RL for making decisions in ATARI games. In contrast to these approaches, our models use deep reinforcement learning to handle high-dimensional sensory inputs.

Recently, methods that integrate deep learning methods with RL have shown promising results. \cite{mnih2015human} propose deep Q networks to play ATARI games. \cite{silver2016} propose a new search algorithm based on the integration of Monte-Carlo tree search with deep RL that beats the world champion in the game of Go. \cite{mnih2016asynchronous} propose a deep RL approach, where the parameters of the deep network are updated by multiple asynchronous copies of the agent in the environment. \cite{levine16} use a deep RL approach to directly map the raw images into torques at robot motors. 
Our work deals with much more complex inputs than ATARI games, or images taken in a lab setting with a constrained background. Additionally, our method is generalizable to new scenes and new targets, while the mentioned methods should be re-trained for a new game, or in case of a change in the game rules.

There have been some effort to develop learning methods that can generalize to different target tasks \cite{policydistallation,actormimic}. In contrast, our model takes the target goal directly as an input without the need of re-training. 

Recently, physics engines have been used to learn the dynamics of real-world scenes from images \cite{wu15, mottaghi16a, mottaghi16}. In this work, we show that a model that is trained in simulation can be generalized to real-world scenarios.

\section{AI2-THOR FRAMEWORK}

To train and evaluate our model, we require a framework for performing actions and perceiving their outcomes in a 3D environment. Integrating our model with different types of environments is a main requirement for generalization of our model. Hence, the framework should have a plug-n-play architecture such that different types of scenes can be easily incorporated. Additionally, the framework should have a detailed model of the physics of the scene so the movements and object interactions are properly represented. 

\begin{figure}[t]
\centering
  \includegraphics[width=1.0\linewidth]{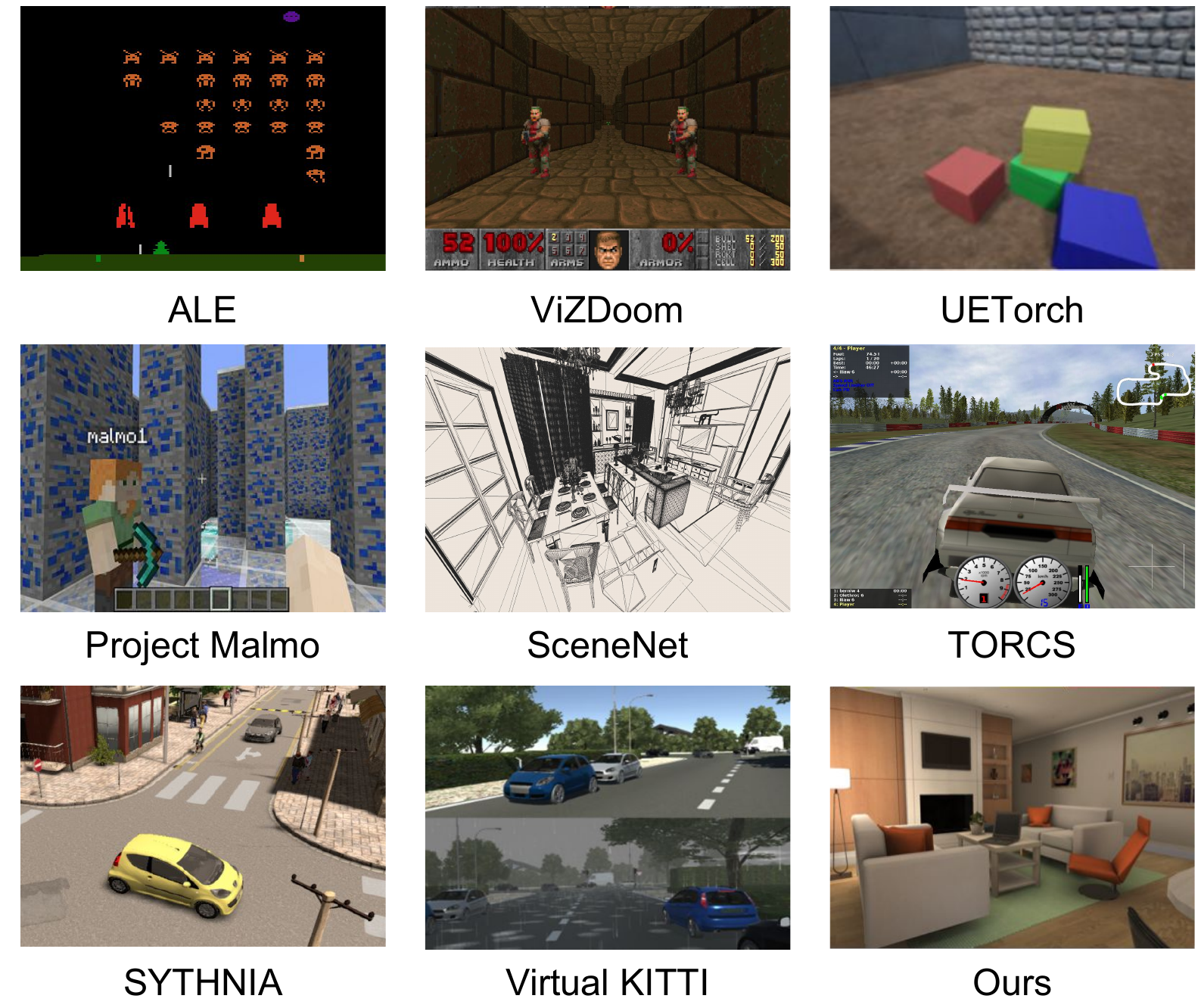}
\caption{Screenshots of our framework and other simulated learning frameworks: ALE~\cite{ALE}, ViZDoom~\cite{ViZDoom}, UETorch~\cite{UETorch}, Project Malmo~\cite{malmo}, SceneNet~\cite{SceneNet}, TORCS~\cite{TORCS}, SYNTHIA~\cite{synthia}, Virtual KITTI~\cite{VirtualKITTI16}.}
\vspace{-0.4cm}
\label{fig:compare}
\end{figure}

For this purpose, we propose The House Of inteRactions (AI2-THOR) framework, which is designed by integrating a physics engine (Unity 3D)
\footnote{http://unity3d.com/}
with a deep learning framework (Tensorflow~\cite{abadi2016tensorflow}). The general idea is that the rendered images of the physics engine are streamed to the deep learning framework, and the deep learning framework issues a control command based on the visual input and sends it back to the agent in the physics engine. Similar frameworks have been proposed by \cite{ALE, ViZDoom, TORCS, malmo, UETorch}, but the main advantages of our framework are as follows: (1) The physics engine and the deep learning framework directly communicate (in contrast to separating the physics engine from the controller as in \cite{mottaghi16}). Direct communication is important since the feedback from the environment can be immediately used for online decision making. (2) We tried to mimic the appearance distribution of the real-world images as closely as possible. For example, \cite{ALE} work on Atari games, which are 2D environments and limited in terms of appearance or \cite{SceneNet} is a collection of synthetic scenes that are non-photo-realistic and do not follow the distribution of real-world scenes in terms of lighting, object appearance, textures, and background clutter, etc. This is important for enabling us to generalize to real-world images.

To create indoor scenes for our framework, we provided reference images to artists to create a 3D scene with the texture and lighting similar to the image. So far we have 32 scenes that belong to 4 common scene types in a household environment: kitchen, living room, bedroom, and bathroom. On average, each scene contains 68 object instances.

\begin{figure}[t]
\centering
  \includegraphics[width=0.9\linewidth]{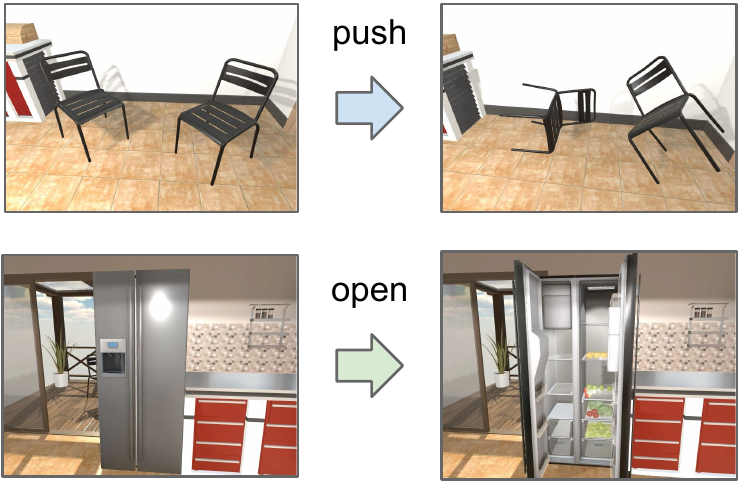}
\caption{Our framework provides a rich interaction platform for AI agents. It enables physical interactions, such as pushing or moving objects (the first row), as well as object interactions, such as changing the state of objects (the second row).}
\vspace{-0.4cm}
\label{fig:interaction}
\end{figure}

The advantage of using a physics engine for modeling the world is that it is highly scalable (training a robot in real houses is not easily scalable). Furthermore, training the models can be performed cheaper and safer (e.g., the actions of the robot might damage objects). One main drawback of using synthetic scenes is that the details of the real world are under-modeled. However, recent advances in the graphics community make it possible to have a rich representation of the real-world appearance and physics, narrowing the discrepancy between real world and simulation. Fig.~\ref{fig:compare} provides a qualitative comparison between a scene in our framework and example scenes in other frameworks and datasets. As shown, our scenes better mimic the appearance properties of real world scenes. In this work, we focus on navigation, but the framework can be used for more fine-grained physical interactions, such as applying a force, grasping, or object manipulations such as opening and closing a microwave. Fig.~\ref{fig:interaction} shows a few examples of high-level interactions. We will provide Python APIs with our framework for an AI agent to interact with the 3D scenes.

\section{TARGET-DRIVEN NAVIGATION MODEL}

In this section, we first define our formulation for target-driven visual navigation. Then we describe our deep siamese actor-critic network for this task.

\subsection{Problem Statement}
Our goal is to find the minimum length sequence of actions that move an agent from its current location to a target that is specified by an RGB image. We develop a deep reinforcement learning model that takes as input an RGB image of the current observation and another RGB image of the target. The output of the model is an action in 3D such as move forward or turn right. Note that the model learns a mapping from the 2D image to an action in the 3D space. 

\subsection{Problem Formulation}

  Vision-based robot navigation requires a mapping from sensory signals to motion commands. Previous work on Reinforcement Learning typically do not consider  high-dimensional perceptual inputs~\cite{kober2013reinforcement}. 
  Recent deep reinforcement learning (DRL) models~\cite{mnih2015human} provide an end-to-end learning framework for transforming pixel information into actions. 
  However, DRL has largely focused on learning goal-specific models that tackle individual tasks in isolation. This training setup is rather inflexible to changes in task goals. For instance, as pointed out by Lake et al.~\cite{lake2016building}, changing the rule of the game would have devastating performance impact on DRL-based Go-playing systems~\cite{silver2016}. Such limitation roots from the fact that standard DRL models~\cite{mnih2015human,mnih2016asynchronous} aim at finding a direct mapping (represented by a deep neural network~$\pi$) from state representations~$s$ to policy~$\pi(s)$. In such cases, the goal is hardcoded in neural network parameters. Thus, changes in goals would require to update the network parameters in accordance.

  Such limitation is especially problematic for mobile robot navigation. When applying DRL to the multiple navigation targets, the network should be re-trained for each target. In practice, it is prohibitive to exhaust every target in a scene. This is the problem caused by a lack of generalization -- i.e., we would have to re-train a new model when incorporating new targets. Therefore, it is preferable to have a single navigation model, which learns to navigate to new targets without re-training. To achieve this, we specify the task objective (i.e., navigation destination) as inputs to the model, instead of implanting the target in the model parameters. We refer to this problem as \textit{target-driven visual navigation}. Formally, the learning objective of a target-driven model is to learn a stochastic policy function $\pi$ which takes two inputs, a representation of current state $s_t$ and a representation of target $g$ and produces a probability distribution over the action space $\pi(s_t, g)$. For testing, a mobile robot keeps taking actions drawn from the policy distribution until reaching the destination. This way, actions are conditioned on both states and targets. Hence, no re-training for new targets is required. 

\subsection{Learning Setup}
\label{sec:learning_setup}
Before introducing our model, we first describe the key ingredients of the reinforcement learning setup: action space, observations and goals, and reward design.
\vspace{1mm}

\subsubsection{Action space} Real-world mobile robots have to deal with low-level mechanics. However, such mechanical details make the learning significantly more challenging. A common approach is to learn at a certain level of abstraction, where the underlying physics is handled by a lower-level controller (e.g., 3D physical engine). We train our model with command-level actions. For our visual navigation tasks, we consider four actions: moving forward, moving backward, turning left, and turning right. We use a constant step length (0.5 meters) and turning angle (90 degree). This essentially discretizes the scene space into a \emph{grid-world} representation. To model uncertainty in real-world system dynamics, we add a Gaussian noise to steps $\mathcal{N}(0, 0.01)$ and turns $\mathcal{N}(0, 1.0)$ at each location. 
\vspace{1mm}

\subsubsection{Observations and Goals} Both observations and goals are images taken by the agent's RGB camera in its first-person view. The benefit of using images as goal descriptions is the flexibility for specifying new targets. Given a target image, the task objective is to navigate to the location and viewpoint where the target image is taken.
\vspace{1mm}

\subsubsection{Reward design} We focus on minimizing the trajectory length to the navigation targets. Other factors such as energy efficiency could be considered instead. Therefore, we only provide a goal-reaching reward ($10.0$) upon task completion. To encourage shorter trajectories, we add a small time penalty (-0.01) as immediate reward.

\begin{figure}[t!]
\begin{center}
\includegraphics[width=1.\linewidth]{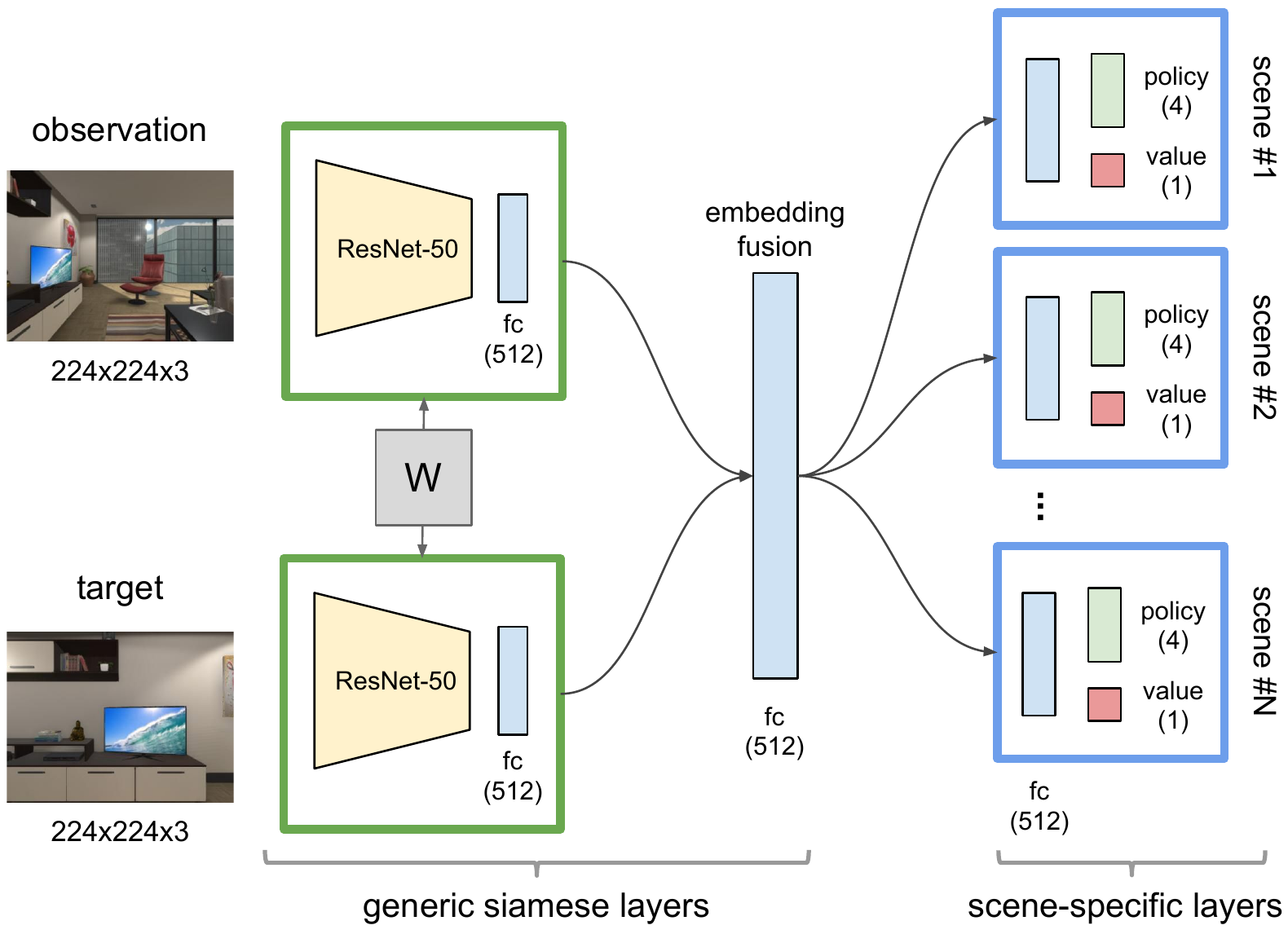}
\caption{Network architecture of our deep siamese actor-critic model. The numbers in parentheses show the output dimensions. Layer parameters in the green squares are shared. The ResNet-50 layers (yellow) are pre-trained on ImageNet and fixed during training.}
\vspace{-0.4cm}
\label{fig:net_arch}
\end{center}
\end{figure}

\subsection{Model}
\label{sec:model}
  
  We focus on learning the target-driven policy function $\pi$ via deep reinforcement learning. We design a new deep neural network as a non-linear function approximator for $\pi$, where action $a$ at time $t$ can be drawn by:
\begin{equation}
a\sim \pi(s_t, g\,|\,\mathbf{u})
\end{equation}
where $\mathbf{u}$ are the model parameters, $s_t$ is the image of the current observation, and $g$ is the image of the navigation target. When target $g$ belongs to a finite discrete set, $\pi$ can be seen as a mixture model, where $g$ indexes the right set of parameters for each goal. However, the number of real-world goals is often countless (due to many different locations or highly variable object appearances). Thus, it is preferable to learn a projection that transforms the goals into an embedding space. Such projection enables knowledge transfer across this embedding space, and therefore allows the model to generalize to new targets. 

Navigational decisions demand an understanding of the relative spatial positions between the current locations and the target locations, as well as a holistic sense of scene layout. We develop a new deep siamese actor-critic network to capture such intuitions. Fig.~\ref{fig:net_arch} illustrates our model for the target-driven navigation tasks. Overall, the inputs to the network are two images that represent the agent's current observation and the target.  Our approach to reasoning about the spatial arrangement between the current location and the target is to project them into the same embedding space, where their geometric relations are preserved. Deep siamese networks are a type of two-stream neural network models for discriminative embedding learning~\cite{chopra2005learning}. We use two streams of weight-shared siamese layers to transform the current state and the target into the same embedding space. Information from both embeddings is fused to form a joint representation. This joint representation is passed through \emph{scene-specific} layers (refer to Fig.~\ref{fig:net_arch}). The intention to have scene-specific layers is to capture the special characteristics (e.g., room layouts and object arrangements) of a scene that are crucial for the navigation tasks. Finally, the model generates policy and value outputs similar to the advantage actor-critic models~\cite{mnih2016asynchronous}. In this model, targets across all scenes share the same generic siamese layers, and all targets within a scene share the same scene-specific layer. This makes the model better generalize across targets and across scenes.

\subsection{Training Protocol}
\label{sec:training_protocol}

Traditional RL models learn for individual tasks in separation, resulting in the inflexibility with respect to goal changes. As our deep siamese actor-critic network shares parameters across different tasks, it can benefit from learning with multiple goals simultaneously. A3C~\cite{mnih2016asynchronous} is a type of reinforcement learning model that learns by running multiple copies of training threads in parallel and updates a shared set of model parameters in an asynchronous manner. It has been shown that these parallel training threads stabilize each other, achieving the state-of-the-art performance in the video-game domain. We use a similar training protocol as A3C. However, rather than running copies of a single game, each thread runs with a different navigation target. Thus, gradients are backpropagated from the actor-critic outputs back to the lower-level layers.
The scene-specific layers are updated by gradients from the navigation tasks within the scene, and the generic siamese layers are updated by all targets.

\subsection{Network Architectures}

The bottom part of the siamese layers are ImageNet-pretrained ResNet-50~\cite{resnet} layers (truncated the softmax layer) that produce $2048$-d features on a $224\times224\times3$ RGB image. We freeze these ResNet parameters during training. We stack 4 history frames as state inputs to account for the agent's previous motions. The output vectors from both streams are projected into the $512$-d embedding space. The fusion layer takes a 1024-d concatenated embedding of the state and the target, generating a 512-d joint representation. This vector is passed through two fully-connected scene-specific layers, producing 4 policy outputs (i.e., probability over actions) and a single value output. We train this network with a shared RMSProp optimizer of learning rate $7\times10^{-4}$.

\section{EXPERIMENTS}

Our main objective for target-driven navigation is to find the shortest trajectories from the current location to the target. In this section, we first evaluate our model with baseline navigation models that are based on heuristics and standard deep RL models. One major advantage of our proposed model is the ability to generalize to new scenes and new targets. We conduct two additional experiments to evaluate the ability of our model to transfer knowledge across targets and across scenes. Also, we show an extension of our model to continuous space. Lastly, we demonstrate the performance of our model in a complex real setting using a real robot.

\subsection{Navigation Results}

We implement our models in Tensorflow~\cite{abadi2016tensorflow} and train them on an Nvidia GeForce GTX Titan X GPU. We follow the training protocol described in Sec.~\ref{sec:training_protocol} to train our deep siamese actor-critic model (see Fig.~\ref{fig:net_arch}) with 100 threads, each thread learns for a different target. It takes around 1.25 hours to pass through one million training frames across all threads.
We report the performance as the average number of steps (i.e., average trajectory length) it takes to reach a target from a random starting point. The navigation performance is reported on 100 different goals randomly sampled from 20 indoor scenes in our dataset. We compare our final model with heuristic strategies, standard deep RL models, and variations of our model.
The models we compare are:

\begin{enumerate}
\item \textbf{Random walk} is the simplest heuristic for navigation. In this baseline model, the agent randomly draws one out of four actions at each step.
\item \textbf{Shortest Path} provides an upper-bound performance for our navigation model. As we discretize the walking space by a constant step length (see Sec.~\ref{sec:learning_setup}), we can compute the shortest paths from the starting locations to the target locations. Note that for computing the shortest path, we have access to the full map of the environment, while the input to our system is just an RGB image.

\item \textbf{A3C}~\cite{mnih2016asynchronous} is an asynchronous advantage actor-critic model that achieves the state-of-the-art results in Atari games. Empirical results show that using more threads improves the data efficiency during training. We thus evaluate A3C model in two setups, where we use 1 thread and 4 threads to train for each target.

\item \textbf{One-step Q}~\cite{mnih2016asynchronous} is an asynchronous variant of deep Q-network~\cite{mnih2015human}.
\item \textbf{Target-driven single branch} is a variation of our deep siamese model that does not have scene-specific branches. In this case, all targets will use and update the same scene-specific parameters, including two FC layers and the policy/value output layers.
\item \textbf{Target-driven final} is our deep siamese actor-critic model introduced in Sec.~\ref{sec:model}.
\end{enumerate}

\begin{figure}[t]
\begin{center}
\includegraphics[width=1.0\linewidth]{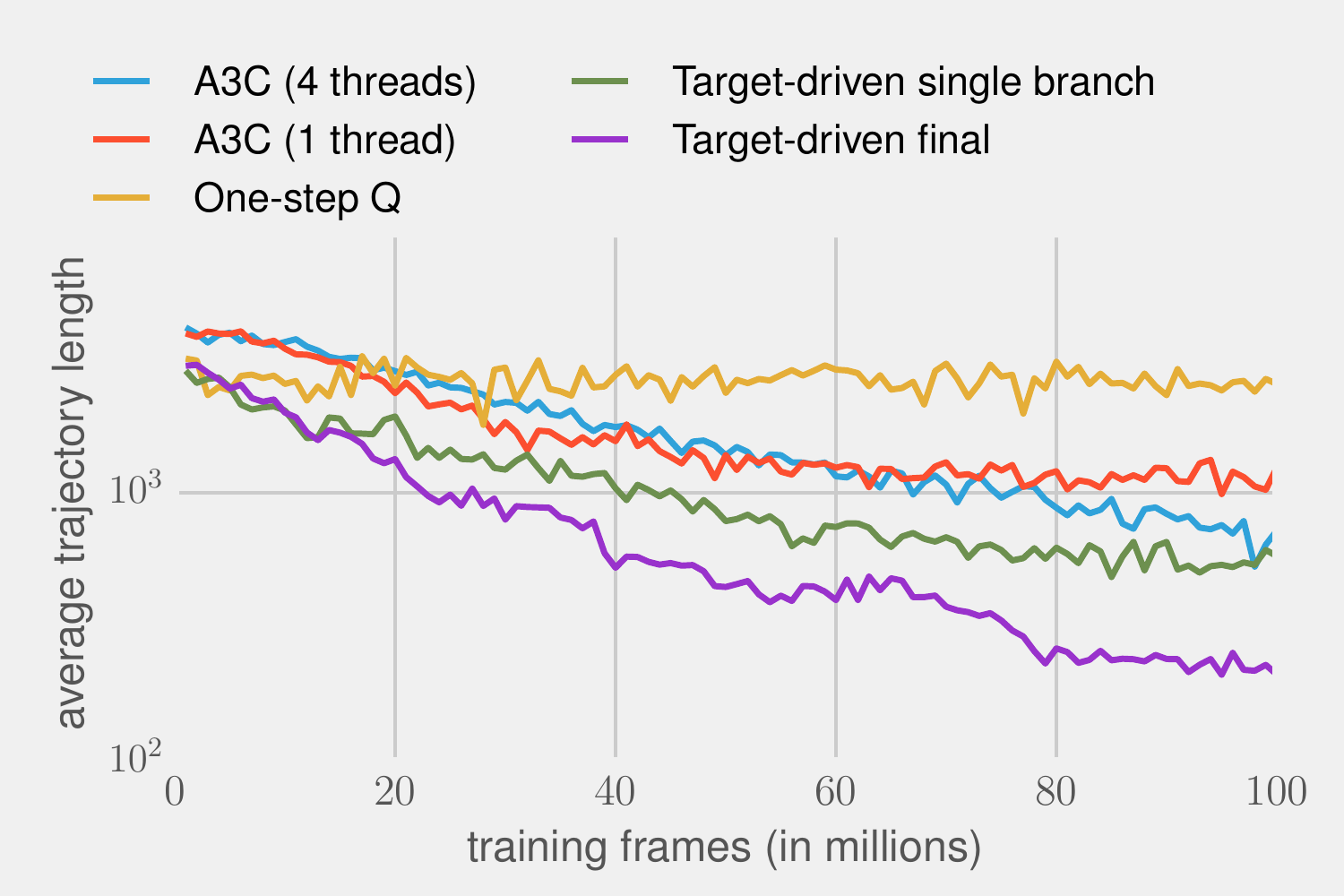}
\caption{\textbf{Data efficiency of training.} Our model learns better navigation policies compared to the state-of-the-art A3C methods~\cite{mnih2016asynchronous} after 100M training frames.}
\vspace{-0.4cm}
\label{fig:method_comparison}
\end{center}
\end{figure}

For all learning models, we report their performance after being trained with 100M frames (across all threads). The performance is measured by the average trajectory length (i.e., number of steps taken) over all targets. An episode ends when either the agent reaches the target, or after it takes 10,000 steps. For each target, we randomly initialize the agent's starting locations, and evaluate 10 episodes. The results are listed in Table~\ref{table:results}.

\begin{table}[htp]
\caption{Performance of Target-driven Methods and Baselines}
\begin{center}
\begin{tabular}{|l|l|c|}
\hline
Type & Method & Avg. Trajectory Length\\
\hline
\hline
Heuristic & Random walk & 2744.3\\
 & Shortest path & 17.6\\
\hline
Purpose-built RL & One-step Q & 2539.2\\
 & A3C (1 thread) & 1241.3\\
 & A3C (4 threads) & 723.5\\
\hline
Target-driven RL & Single branch & 581.6\\
\multicolumn{1}{|l|}{(Ours)} & Final & \textbf{210.7}\\
\hline
\end{tabular}
\end{center}
\vspace{-0.4cm}
\label{table:results}
\end{table}%

We analyze the data efficiency of learning with the learning curves in Fig.~\ref{fig:method_comparison}. Q-learning suffers from slow convergence. A3C performs better than Q-learning; plus, increasing the number of actor-learning threads per target from 1 to 4 improves learning efficiency. Our proposed target-driven navigation model significantly outperforms standard deep RL models when it uses 100M frames for training. We hypothesize that this is because both the weight sharing scheme across targets and the asynchronous training protocol facilitate learning generalizable knowledge. In contrast, purpose-built RL models are less data-efficient, as there is no straightforward mechanism to share information across different scenes or targets. The average trajectory length of the final model is three times shorter than the one of the \emph{single branch} model. It justifies the use of scene-specific layers, as it captures particular characteristics of a scene that may vary across scene instances.

\begin{figure}[htbp]
\begin{center}
\includegraphics[width=1.0\linewidth]{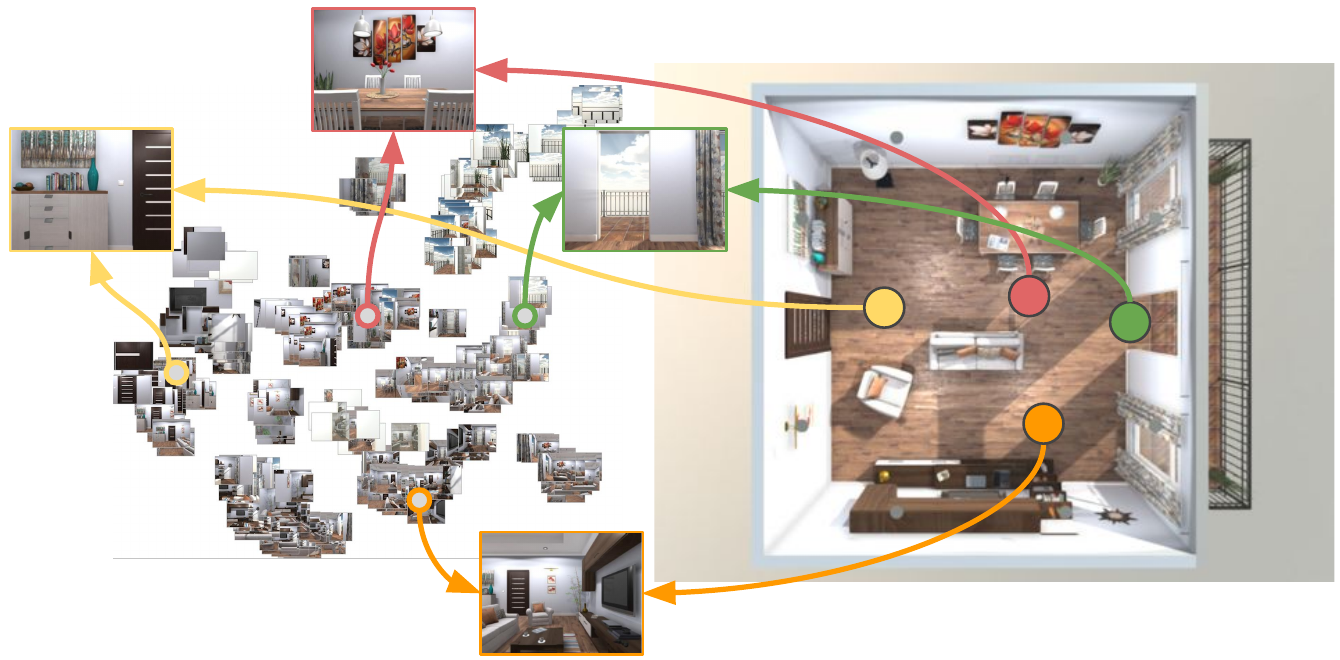}
\caption{\textbf{t-SNE embeddings of observations in a living room scene.} We highlight four observation examples in the projected 2D space and their corresponding locations in the scene (bird's-eye view on the right). This figure shows that our model has learned observation embeddings while preserving their relative spatial layout.}
\vspace{-0.4cm}
\label{fig:tsne_embedding}
\end{center}
\end{figure}

To understand what the model learns, we examine the embeddings learned by generic siamese layers. Fig.~\ref{fig:tsne_embedding} shows t-SNE visualization~\cite{maaten2008visualizing} of embedding vectors computed from observations at different locations at four different orientations. We observe notable spatial correspondence between the spatial arrangement of these embedding vectors and their corresponding t-SNE projections. We therefore hypothesize that the model learns to project observation images into the embedding space while preserving their spatial configuration. To validate this hypothesis, we compare the distance of pairwise projected embeddings and the distance of their corresponding scene coordinates. The Pearson correlation coefficient is 0.62 with p-value less than 0.001, indicating that the embedding space preserves information of the original locations of observations. This means that the model learns a rough map of the environment and has the capability of localization with respect to this map.

\subsection{Generalization Across Targets}

In addition to the data-efficient learning of our target-driven models, our model has the built-in ability to generalize, which is a significant advantage over the purpose-built baseline models. We evaluate its generalization ability in two dimensions: 1. generalizing to new targets within one scene, and 2. generalizing to new scenes. We focus on generalization across targets in this section, and explain scene generalization in Sec.~\ref{sec:generalization_across_scenes}.

\begin{figure}[htbp]
\begin{center}
\includegraphics[width=1.0\linewidth]{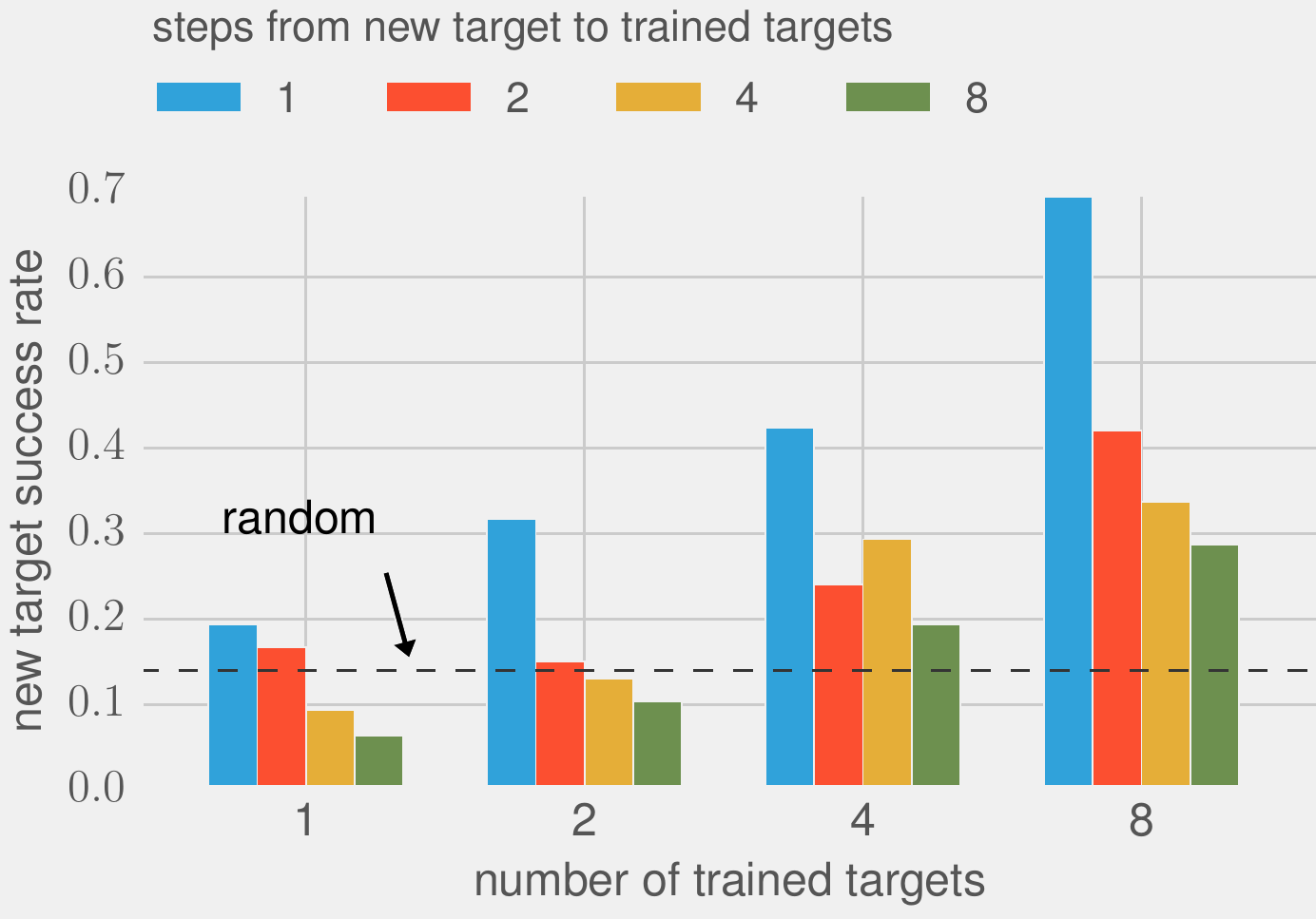}
\caption{\textbf{Target generalization.} Each histogram group reports the success rate of navigation to new targets with certain number of trained targets. The four bars in each group indicate the impact of adjacency between the trained and new targets on generalization performance.}
\vspace{-0.4cm}
\label{fig:target_generalization}
\end{center}
\end{figure}

We test the model to navigate to new targets that are excluded from training. We take 10 of the largest scenes in our dataset, each having around 15 targets. We gradually increase the number of trained targets (from 1, 2, 4 to 8) using our target-driven model. All models are trained with 20M frames. During testing, we run 100 episodes for each of 10 new targets. 
These new targets are randomly chosen from a set of locations that have a constant distance (1, 2, 4 and 8 steps) from the nearest trained targets. The results are illustrated in Fig.~\ref{fig:target_generalization}. We use \emph{success rate} (percentage of trajectories shorter than 500 steps) to measure the performance. We choose this metric due to the bipolar behavior of our model on new targets -- it either reaches the new targets quickly, or fails completely. Thus, this metric is more effective than average trajectory lengths.
In Fig.~\ref{fig:target_generalization}, we observe a consistent trend of increasing success rate, as we increase the number of trained targets (x-axis). Inside each histogram group, the success rate positively correlates with adjacency between trained and new targets. It indicates that the model has a clearer understanding of nearby regions around the trained targets than distant locations.

\subsection{Generalization Across Scenes}
\label{sec:generalization_across_scenes}
\begin{figure}[htbp]
\begin{center}
\includegraphics[width=1.0\linewidth]{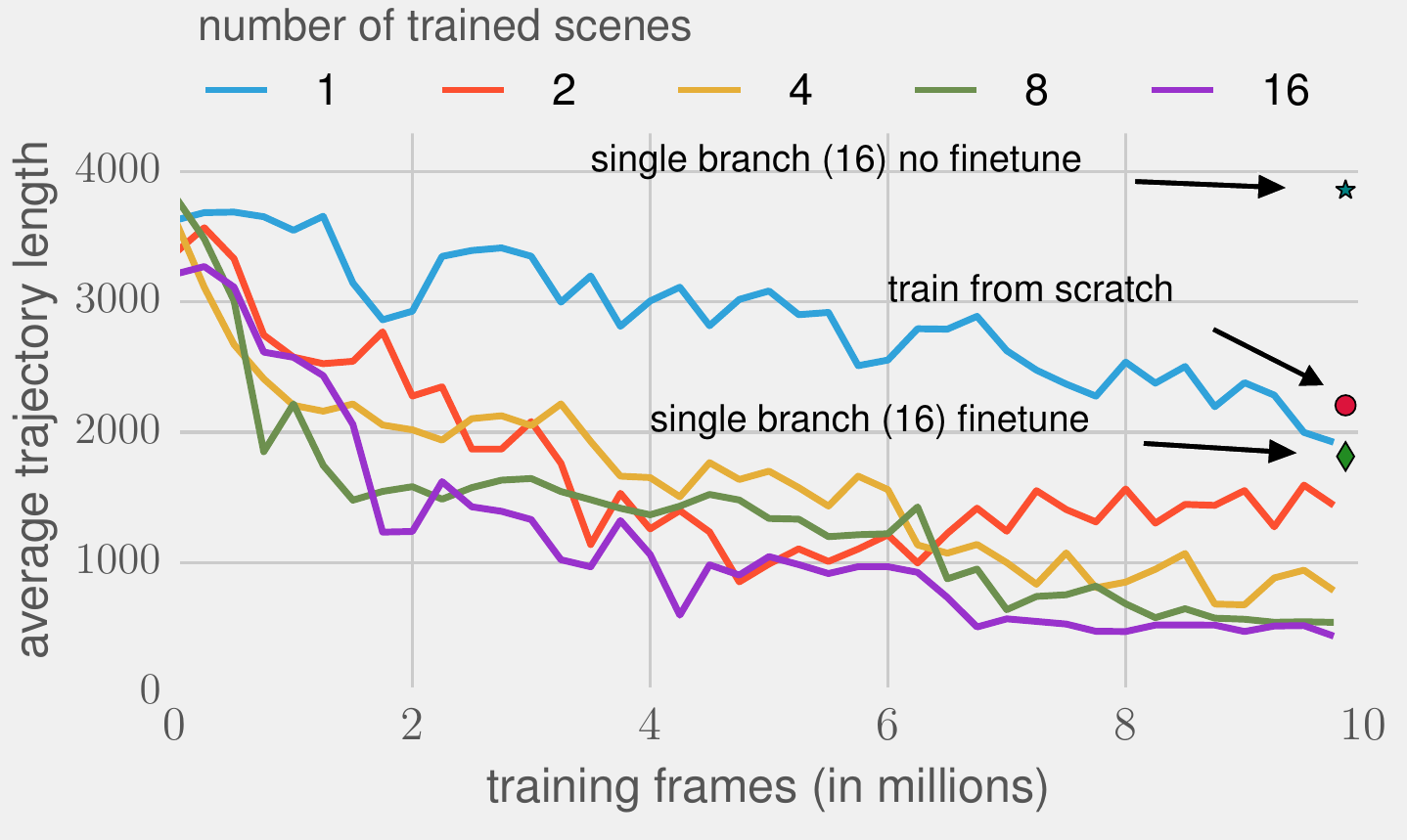}
\caption{\textbf{Scene generalization.} We compare the data efficiency for adapting trained navigation models to unseen scenes. As the number of trained scene instances increases, fine-tuning the scene-specific layers becomes faster.}
\vspace{-0.4cm}
\label{fig:scene_generalization}
\end{center}
\end{figure}

We further evaluate our model's ability to generalize across scenes. As the generic siamese layers are shared over all scenes, we examine the possibility of transferring knowledge from these layers to new scenes. Furthermore, we study how the number of trained scenes would influence the transferability of generic layer parameters. We gradually increase the number of trained scenes from 1 to 16, and test on 4 unseen scenes. We select 5 random targets from each scene for training and testing. To adapt to unseen scenes, we train the scene-specific layers while fixing generic siamese layers. Fig.~\ref{fig:scene_generalization} shows the results. We observe faster convergence as the number of trained scenes grows. Compared to training from scratch, transferring generic layers significantly improves data efficiency for learning in new environments. We also evaluate the \emph{single branch} model in the same setup. As the \emph{single branch} model includes a single scene-specific layer, we can apply a trained model (trained on 16 scenes) to new scenes without extra training. However, it results in worse performance than chance, indicating the importance of adapting scene-specific layers. The \emph{single branch} model leads to slightly faster convergence than training from scratch, yet far slower than our final model.

\subsection{Continuous Space}

The space discretization eliminates the need for handling complex system dynamics, such as noise in motor control. In this section, we show empirical results that the same learning model is capable of coping with more challenging continuous space. 

To illustrate this, we train the same target-driven model for a door-finding task in a large living room scene, where the goal is to arrive at the balcony through a door. We use the same 4 actions as before (see Sec.~\ref{sec:learning_setup}); however, the agent's moves and turns are controlled by the physics engine. In this case, the method should explicitly handle forces and collisions, as the agent may be stopped by obstacles or slide along heavy objects.  Although this setting requires significantly more training frames (around 50M) to train for a single target, the same model learns to reach the door in average 15 steps, whereas random agents take 719 steps on average. We provide sample test episodes in the supplementary video.

\subsection{Robot Experiment}
\begin{figure}[t]
\begin{center}
\includegraphics[width=1.0\linewidth]{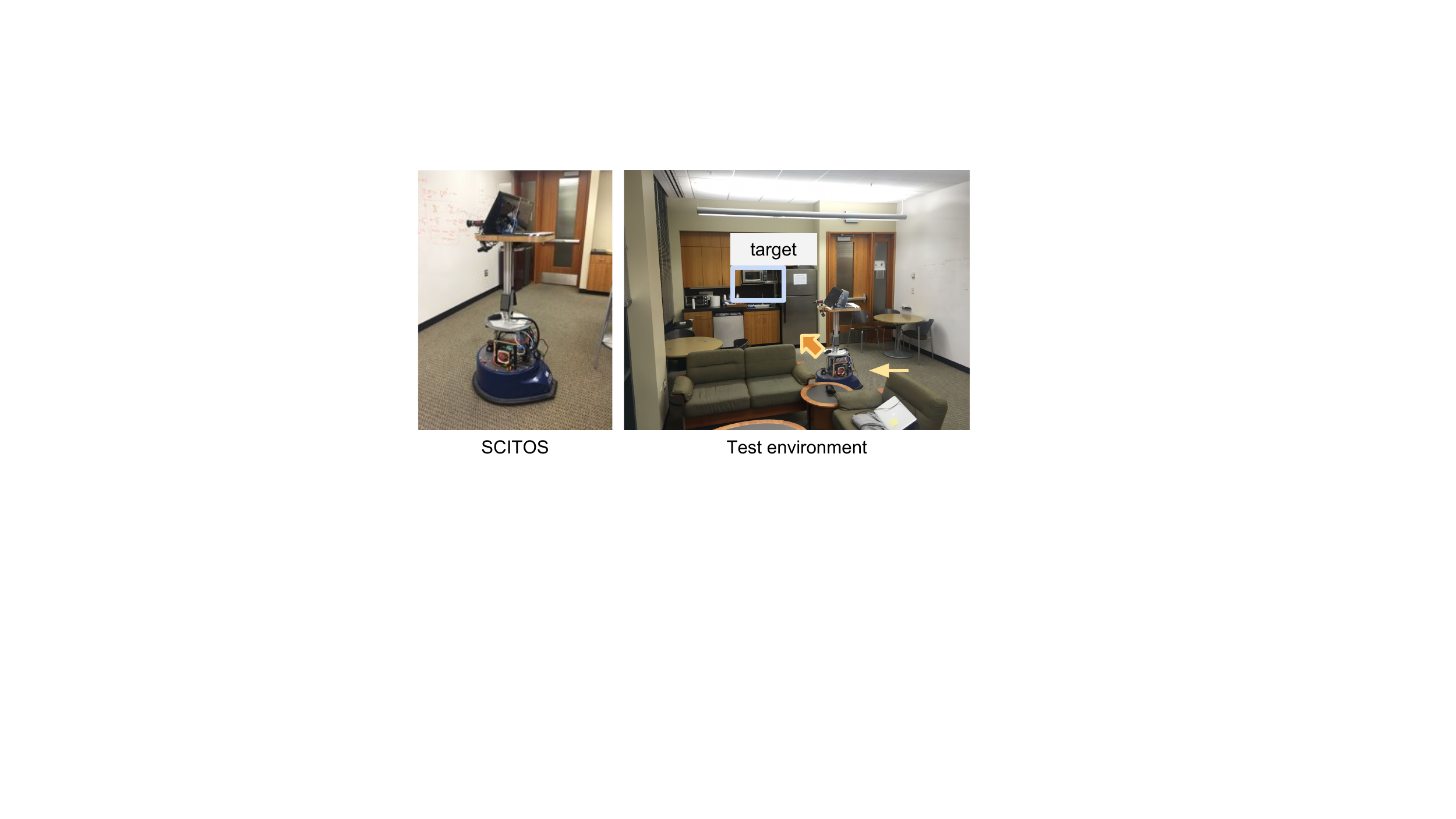}
\caption{\textbf{Robot experiment setup.} Our experiments are conducted on a SCITOS mobile robot. On the left, we show a picture of the SCITOS robot. On the right, we show the test environment and one target (microwave) that we have used for evaluation.}
\vspace{-0.4cm}
\label{fig:realrobot}
\end{center}
\end{figure}

To validate the generalization of our method to real world settings, we perform an experiment by using a SCITOS mobile robot modified by~\cite{chung2016iros} (see Fig.~\ref{fig:realrobot}). We train our model in three different settings: 1) training on real images from scratch; 2) training only scene-specific layers while freezing generic layer parameters trained on 20 simulated scenes; and 3) training scene-specific layers and fine-tuning generic layer parameters.

We train our model (with backward action disabled) on 28 discrete locations in the scene, which are roughly 30 inches apart from each other in each dimension. At each location, the robot takes 4 RGB images (90 degrees apart) using its head camera. During testing, the robot moves and turns based on the model's predictions. We evaluate the robot with two targets in the room: \emph{door} and \emph{microwave}. Although the model is trained on a discretized space, it exhibits robustness towards random starting points, noisy dynamics, varying step lengths, changes in illumination and object layouts, etc. Example test episodes are provided in the supplementary video. All three setups converge to nearly-optimal policy due to the small scale of the real scene. However, we find that transferring and fine-tuning parameters from simulation to real data offers the fastest convergence out of these three setups (44\% faster than training from scratch). This provides supportive evidence on the value of simulations in learning real-world interactions and shows the possibility of generalization from simulation to real images using a small amount of fine-tuning.

\section{CONCLUSIONS}

We proposed a deep reinforcement learning (DRL) framework for target-driven visual navigation. The state-of-the-art DRL methods are typically applied to video games and environments that do not mimic the distribution of natural images. This work is a step towards more realistic settings.

The state-of-the-art DRL methods have some limitations that prevent them from being applied to realistic settings. In this paper, we addressed some of these limitations. We addressed generalization across scenes and targets, improved data efficiency compared to the state-of-the-art DRL methods, and provided AI2-THOR framework that enables inexpensive and efficient collection of action and interaction data. 

Our experiments showed that our method generalizes to new targets and scenes that are not used during the end-to-end training of the model. We also showed our method converges with much fewer training samples compared to the state-of-the-art DRL methods. Furthermore, we showed that the method works in both discrete and continuous domains. We also showed that a model that is trained on simulation can be adapted to a real robot with a small amount of fine-tuning. We provided visualizations that show that our DRL method implicitly performs localization and mapping. Finally, our method is end-to-end trainable. Unlike the common visual navigation methods, it
does not require explicit feature matching or 3D reconstruction of the environment.

Our future work includes increasing the number of high-quality 3D scenes in our framework. Additionally, we plan to build models that learn the physical interactions and object manipulations in the framework.

\small{
\section*{Acknowledgements}
We would like to thank Dieter Fox for his helpful comments, Andrzej Pronobis and Yu Xiang for helping us with the robot experiments, and Noah Siegel for his help with creating the video.}

\bibliographystyle{IEEEtran}
\bibliography{egbib}

\end{document}